%% file: arxiv_main.tex
\documentclass[pmlr]{jmlr}

\usepackage{longtable}
\usepackage{booktabs}
\usepackage{multirow}
\usepackage{xcolor}
\usepackage{pgfplots}    
\pgfplotsset{compat=1.18}

\makeatletter
\let\Ginclude@graphics\@org@Ginclude@graphics
\makeatother

\graphicspath{{figures/}}

\jmlrvolume{353,} 
\jmlryear{2026}
\jmlrworkshop{ACML 2026}

\title[Invertible QK Coupling]{Invertible Query-Key Coupling Composes with Attention Mechanisms}

\author{\Name{Barak Gahtan} \Email{barakgahtan@cs.technion.ac.il}\\
\addr Technion -- Israel Institute of Technology, Haifa, Israel
\AND
\Name{Alex M. Bronstein} \Email{bron@cs.technion.ac.il}\\
\addr Technion -- Israel Institute of Technology, Haifa, Israel\\
\addr Institute of Science and Technology Austria, Klosterneuburg, Austria}

\editors{Andy Song, Bo Han and Sarah Erfani}

\begin{document}

\makeatletter
\let \@jmlrpages \@empty
\makeatother

\maketitle

\input{sections/abstract}


\input{sections/introduction}
\input{sections/method}
\input{sections/experiments}
\input{sections/related}
\input{sections/conclusion}

\bibliography{references}

\appendix
\input{sections/appendix}
\input{sections/supp_controls}
\input{sections/supp_overhead}
\input{sections/supp_perseed}
\input{sections/supp_testppl}
\input{sections/supp_analysis}
\input{sections/supp_optstability}
\input{sections/supp_entropy}
\input{sections/supp_hamiltonian}

\end{document}

%% file: sections/abstract.tex
\begin{abstract}
Scaled dot-product attention forms its queries and keys as independent linear projections of the input, so the two never interact before the dot product that produces attention scores.
We study coupled query-key dynamics, a pre-scoring transformation that evolves each token's own query and key jointly through a shared invertible coupling before standard scoring, leaving the interaction across token positions to the dot product as usual.
We realize it as an alternating affine map in the style of real-valued non-volume-preserving flows: the coupling is the identity at initialization, adds a small fraction of parameters per head, and leaves the softmax and the surrounding architecture unchanged.
We therefore place coupling on top of existing attention methods instead of replacing them, and ask whether that composition helps.
On WikiText-103 language modeling, adding coupling to Differential Attention improves on it at both 150M and 455M parameters.
At 455M the gain is significant at sequence length 512 ($p{=}0.003$, six seeds), survives a Bonferroni correction across our family of tests, and replicates on a held-out test split; it also holds across rotary-embedding training lengths of 512, 1024, and 2048, where the 1024 comparison likewise survives the correction.
The same additive direction appears when coupling is added to query-key normalization (significant at 150M) and Multi-Token Attention (directional); neither is significant at 455M, where the Differential Attention composition alone holds.
Matched controls attribute the gain to the joint pre-scoring coupling itself rather than to added capacity, and show that removing the invertibility guarantee preserves the 455M gain yet is far worse than the base method at 150M, so invertibility is what makes the coupling reliable across scales.
On its own, by contrast, coupling lowers perplexity at 60M and 150M (one-sided Welch tests, $p<0.05$, three seeds), but this gain narrows with scale and is not significant at 455M.
We relate the construction to the expressivity of coupling flows, use an associative-recall study to map where coupling helps and where it degrades sharp retrieval, and state the limitations directly, concluding that coupling is most useful in composition with a scoring-stage method rather than as a standalone change.
\end{abstract}

%% file: sections/introduction.tex
\section{Introduction}
\label{sec:intro}

\paragraph{Queries and keys do not interact before scoring.}
The attention mechanism~\citep{vaswani2017attention} is the central computational primitive of the modern Transformer.
A standard attention head forms a query matrix $Q$ and a key matrix $K$ as two independent linear projections of the same input, then scores token pairs through $\mathrm{softmax}(QK^\top/\sqrt{d_k})$.
Because $Q$ and $K$ are produced separately, neither carries any information about the other until the dot product that scores them.
A growing body of work improves attention by changing this scoring stage, either by acting on the scores after the product or by normalizing the projections, while the interaction between $Q$ and $K$ before the product has been studied far less.

\begin{figure}[t]
\centering
\begin{tikzpicture}
\begin{axis}[
  width=0.62\columnwidth, height=0.36\columnwidth,
  xmode=log, log basis x=10,
  xtick={60e6,150e6,455e6}, xticklabels={60M,150M,455M},
  xlabel={Model size (parameters)}, ylabel={Perplexity reduction (\%)},
  ymin=-0.1, ymax=1.45, xmin=45e6, xmax=620e6,
  legend style={at={(0.97,0.97)}, anchor=north east, draw=none, font=\footnotesize, fill=none},
  tick label style={font=\footnotesize}, label style={font=\small},
  ymajorgrids=true, grid style={gray!18},
]
\addplot[color=red!70!black, mark=*, thick, mark size=2.4pt]
  coordinates {(60e6,1.175) (150e6,0.744) (455e6,0.360)};
\addlegendentry{Coupling alone vs.\ standard}
\addplot[color=blue!60!black, mark=square*, thick, mark size=2.4pt]
  coordinates {(150e6,0.828) (455e6,0.513)};
\addlegendentry{Coupling ${+}$ Differential vs.\ Differential}
\addplot[only marks, mark=*, mark options={fill=white, draw=red!70!black}, mark size=2.4pt, forget plot]
  coordinates {(455e6,0.360)};
\end{axis}
\end{tikzpicture}
\caption{The standalone coupling gain narrows with scale (red), while the gain from adding coupling on top of Differential Attention persists at 150M and 455M (blue). Values are relative perplexity reduction on WikiText-103 computed from per-seed perplexities, so they can differ by rounding from the means in Tables~\ref{tab:scaling} and~\ref{tab:compose}; filled markers are significant (one-sided Welch $p<0.05$). The open marker is the standalone coupling at 455M, which is not significant ($p{=}0.19$); the 455M composition gain is significant ($p{=}0.003$).}
\label{fig:axis}
\end{figure}

\paragraph{Two established directions, and a gap.}
Recent methods cluster into two families.
Scoring-stage methods reshape the attention scores after the dot product: Differential Attention subtracts two softmax maps to cancel common-mode noise~\citep{ye2025diff}, and Multi-Token Attention convolves over the score map to mix neighboring query and key positions~\citep{golovneva2025multitoken}.
Projection-level methods normalize or rescale $Q$ and $K$, as in query-key normalization (QK-Norm)~\citep{henry2020query} and fully normalized Transformers~\citep{loshchilov2024ngpt}.
A learned joint transformation of $Q$ and $K$ before the dot product, mapping $(q,k)$ to a new $(q', k')$ pair, remains underexplored. The closest prior forms are the bilinear scoring function $q^\top W k$~\citep{luong2015effective} and additive attention~\citep{bahdanau2015neural}, but both produce a scalar score rather than an evolved $(q', k')$ pair, and neither is designed to compose with scoring-stage methods.
Two obstacles explain the caution.
A pre-scoring change can act as little more than added head capacity, so any gain must be separated from a parameter-count effect.
And pre-scoring gains reported at small scale often shrink as models grow, which leaves open whether such a change is useful at the scale where it would be deployed.

\paragraph{Coupled query-key dynamics.}
We study a pre-scoring transformation that evolves $Q$ and $K$ jointly through a shared, invertible coupling before standard scoring, which we call \emph{coupled query-key dynamics}.
We use \emph{dynamics} to mean the mutual, multi-step evolution of $Q$ and $K$ through the coupling rounds, not a temporal differential equation; the Hamiltonian analogy that first motivated the design is described in the supplementary material.
Our realization is an alternating affine coupling in the style of real-valued non-volume-preserving flows~\citep{dinh2017density}: the head dimension is split, and the two halves alternately scale and shift one another while their roles as query and key swap between steps.
The map is invertible by construction, reduces to the identity at initialization, adds a small fraction of parameters per head, and changes neither the softmax nor anything downstream.
The coupling acts within each token, mapping a token's own $(q,k)$ to a new $(q',k')$; the query at position $i$ and the key at position $j$ still meet only at the dot product, so coupling enriches the per-token query and key representations rather than introducing a pairwise pre-scoring interaction.
Two properties motivate this specific form.
Coupling flows of this kind are universal approximators of invertible smooth maps given sufficient depth~\citep{teshima2020coupling}, whereas a purely additive, volume-preserving coupling, which is where we began, is provably restricted; the affine, non-volume-preserving, alternating form is what removes that restriction, though our fixed, shallow per-head realization is not claimed to inherit the universality guarantee itself.
We therefore treat coupling not as a replacement for the scoring-stage methods above but as a complementary, composable transformation that can be placed on top of them; we use the word axis in the empirical sense that coupling produces a consistent additive direction across scoring-stage and projection-level methods, not as a claim of structural independence.

\paragraph{What we find.}
A standalone coupling lowers WikiText-103 perplexity at 60M and 150M parameters, with seed distributions that do not overlap the standard-attention baseline (one-sided Welch tests, $p<0.05$).
The size of this standalone gain shrinks as the model grows and is not significant at 455M, so we do not claim a standalone improvement that survives scaling.
The result that does survive is compositional (Figure~\ref{fig:axis}).
Adding coupling on top of Differential Attention improves on Differential Attention alone at 150M (three seeds) and at 455M (six seeds, $p{=}0.003$ at 455M), and the same additive direction appears when coupling is placed on top of Multi-Token Attention (not significant) and query-key normalization (significant at 150M).
The 455M compositional gain also holds at rotary-embedding training sequence lengths 512, 1024, and 2048, with relative reductions of $1.0\%$, $1.5\%$, and $2.1\%$; the 1024 comparison survives our Bonferroni correction ($p{=}0.0015$) while the 512 and 2048 comparisons do not, and we read the larger reduction at longer lengths as descriptive (Figure~\ref{fig:seqlen}).
Because one small pre-scoring map improves Differential Attention and moves Multi-Token Attention and query-key normalization in the same direction, and because each composed variant is identical to its base method at initialization, the gain is not naturally explained by a single interacting method.
Two matched controls at 455M sharpen this attribution: added non-invertible pre-scoring capacity reproduces none of the gain, and a non-invertible variant of the coupling itself preserves the 455M gain but is far worse than the base method at 150M, so the gain comes from the joint coupling and invertibility is what makes it reliable across scales (\S\ref{sec:experiments}).
We relate the construction to the expressivity of coupling flows, and we use a multi-query associative-recall task to separate the regime where coupling helps from the regime where it suppresses the sharp retrieval head that the task needs.

\paragraph{Contributions.}
\begin{itemize}
\item We frame the joint pre-scoring transformation of queries and keys as a design axis complementary to scoring-stage attention methods, and give a parameter-light, identity-initialized realization through an alternating affine coupling (\S\ref{sec:method}).
\item We complete the coupling family with a controlled ablation, from standard attention to a one-way affine map to the alternating affine coupling, and show with significance tests that the alternating, non-volume-preserving form is the one that helps; matched capacity and invertibility controls at 455M attribute the compositional gain to the joint coupling, with invertibility acting as the property that keeps it reliable across scales (\S\ref{sec:experiments}).
\item We show that coupling \emph{composes} with established attention methods, which we take as the central claim rather than a standalone win: it improves Differential Attention at 150M and 455M ($p{=}0.003$ across six seeds at 455M, surviving a Bonferroni correction and holding across rotary training lengths 512--2048, with the 1024 comparison surviving the correction), and moves Multi-Token Attention and query-key normalization in the same direction, even though the standalone gain is not significant at 455M (\S\ref{sec:experiments}).
\item We characterize the scope of the mechanism. On associative recall, coupling helps at moderate difficulty and degrades the sharpest retrieval setting, which we report together with the scale limitation rather than omitting it (\S\ref{sec:experiments}).
\end{itemize}

%% file: sections/method.tex
\section{Method}
\label{sec:method}

\paragraph{Preliminaries.}
A standard scaled dot-product attention head~\citep{vaswani2017attention} forms queries, keys, and values as linear projections of the input and computes
\begin{equation}
\mathrm{Attn}(Q, K, V) = \mathrm{softmax}\!\left(\frac{QK^\top}{\sqrt{d_k}}\right) V .
\label{eq:sdpa}
\end{equation}
We separate the head computation into three stages: a \emph{pre-scoring} stage that produces $Q$ and $K$, a \emph{scoring} stage that maps them to logits and through the softmax, and the value aggregation.
Most recent attention variants act on the scoring stage or on the values.
Coupling acts only on the pre-scoring stage.

\paragraph{Coupled query-key dynamics.}
We insert a learned map $\mathcal{C}:(Q,K)\mapsto(\tilde Q,\tilde K)$ that acts per head before the dot product, after which scoring proceeds as in Eq.~\eqref{eq:sdpa} with $\tilde Q,\tilde K$ in place of $Q,K$.
We require $\mathcal{C}$ to be invertible, so it redistributes rather than discards information, and to equal the identity at initialization, so a trained model can recover standard attention; the controls in \S\ref{sec:experiments} support this design choice empirically, showing that a non-invertible variant of the same coupling can match the composition gain at one scale yet fail badly at another, so the guarantee functions as the property that keeps the coupling reliably beneficial.
A pre-scoring map of this kind is distinct from the two pre-scoring operations already in common use: it is learned and joint, unlike the fixed rotation of rotary position encodings, and it mixes the two vectors, unlike the per-vector rescaling of query-key normalization~\citep{henry2020query}.

\paragraph{Alternating affine coupling.}
We realize $\mathcal{C}$ as an alternating affine coupling in the style of real-valued non-volume-preserving flows~\citep{dinh2017density}.
For a head with per-token query $q$ and key $k$ in $\mathbb{R}^{d_k}$, one round applies
\begin{align}
k' &= k \odot \exp\!\big(s_\theta(q)\big) + t_\theta(q), \label{eq:coupleK}\\
q' &= q \odot \exp\!\big(s_\phi(k')\big) + t_\phi(k'), \label{eq:coupleQ}
\end{align}
so the query first conditions an affine transform of the key, then the updated key conditions an affine transform of the query.
Each scale and shift pair $(s,t)$ is a two-layer per-head network with a SiLU nonlinearity whose output layer is initialized to zero, which makes $s=0$ and $t=0$ and hence $(\tilde q,\tilde k)=(q,k)$ at the start of training.
The map is invertible for any value of the networks, and its log-volume change is $\sum s_\theta(q) + \sum s_\phi(k')$, which is nonzero in general.
We apply a small fixed number of rounds.
This construction adds roughly $0.3\%$ parameters at 60M and does not change the $\mathcal{O}(n^2 d_k)$ cost of attention in sequence length.

\paragraph{Why this form.}
Coupling flows that alternate which coordinates are transformed are universal approximators of invertible smooth maps given sufficient depth~\citep{teshima2020coupling}, so Eqs.~\eqref{eq:coupleK}--\eqref{eq:coupleQ} can in principle express a wide family of pre-scoring reparametrizations of the query-key geometry, such as axis-aligned rescalings that sharpen or broaden selectivity.
We invoke this only to motivate the alternating affine form over a volume-preserving one; it is not evidence that the shallow, fixed-depth per-head networks we train realize that full family, so the ablation, not the theorem, is what supports our claims.
A purely additive coupling, with the scale fixed to one, is volume-preserving and provably more restricted; we began from such an additive, symplectic update inspired by Hamiltonian dynamics and adopt the affine, non-volume-preserving form because the additive one cannot represent the same family of maps, with that origin described in the supplementary material.
The controlled ladder in \S\ref{sec:experiments} confirms that the alternating affine step is the one that carries the significant gain.

\paragraph{Composition with scoring-stage methods.}
Because coupling produces a drop-in replacement for $(Q,K)$ and leaves the softmax untouched, it composes with any method that modifies the scoring stage.
We use three composed variants.
\textbf{Coupling${+}$Differential Attention} applies Eqs.~\eqref{eq:coupleK}--\eqref{eq:coupleQ} and then the dual-softmax scoring of Differential Attention~\citep{ye2025diff}.
\textbf{Coupling${+}$Multi-Token Attention} applies the coupling and then the score-map convolution of Multi-Token Attention~\citep{golovneva2025multitoken}.
\textbf{Coupling${+}$QK-Norm} applies the coupling and then query-key normalization~\citep{henry2020query}, a projection-level method we include as a third, independent mechanism.
By the zero initialization of the coupling networks, each composed variant reduces \emph{exactly} to its base method at initialization, so any change in perplexity is attributable to the added coupling rather than to a different starting point.

\paragraph{Baselines and protocol.}
We compare against standard attention and the three base methods above used alone, and we report grouped-query attention~\citep{ainslie2023gqa} as a parameter-budget reference rather than a quality baseline (supplementary material).
We do not include Elliptical Attention as a standalone baseline: it replaces the dot product with a learned quadratic form, so it is not a drop-in scoring-stage method that coupling composes with, and in our setup its standalone perplexity did not improve on standard attention.
All methods share the same optimizer, schedule, and token budget, and we compare at matched optimization steps; the coupling adds about $0.3\%$ parameters and a measured wall-clock overhead of roughly $10\%$ at 60M on a single H100, shrinking to about $4$ to $7\%$ at 455M, benchmarked in the supplementary material.

%% file: sections/experiments.tex
\section{Experiments} \label{sec:experiments}

\paragraph{Setup.}
We train decoder-only Transformers on WikiText-103~\citep{merity2017pointer} at three sizes: 60M, 150M ($d{=}768$, $H{=}12$, $L{=}12$), and 455M ($d{=}1024$, $H{=}16$, $L{=}24$) parameters. Each model uses learned positional embeddings, SwiGLU feed-forward layers, and RMSNorm. Training runs for 30K steps with AdamW, a cosine learning rate, sequence length 512, and the GPT-2 byte-pair vocabulary.
All attention variants use the same harness, token budget, and step count; the only change is the attention head.
We run three seeds at each size for the headline comparisons (six for the 455M Differential Attention composition, our primary claim), and report perplexity as the exponential of the best validation loss, with mean and standard deviation; cross-seed variance is examined further in the supplementary material.
We assess significance with one-sided Welch $t$-tests and report Cohen's $d$.
We additionally measure generalization on a second corpus (FineWeb-Edu) and zero-shot transfer to LAMBADA and HellaSwag.
All runs use H100 GPUs; full model configurations and per-variant hyperparameters are in Appendix~\ref{app:setup}.

\paragraph{The coupling family.}
\label{sec:ablation}
Table~\ref{tab:family} completes the family from standard attention to a one-way affine map to the alternating affine coupling.
Each step lowers perplexity; the move from the one-way map to the alternating coupling is the strongly significant step at 60M ($p{=}0.006$, $d{=}4.4$), while the intermediate steps are weaker or marginal (the full ladder with $p$-values is in Appendix~\ref{app:stats}).
This ordering matches the expressivity argument in \S\ref{sec:method}: a purely additive, volume-preserving coupling is provably restricted, whereas the affine, non-volume-preserving, alternating form is a universal approximator of invertible maps, and it is the alternating step that carries the significant gain.

\begin{table}[t]
\centering
\caption{The coupling family on WikiText-103, three seeds, mean$\pm$std perplexity. Each step lowers perplexity; the alternating step is the strongly significant one at 60M ($p{=}0.006$), while the intermediate steps are weaker or marginal (full ladder in Appendix~\ref{app:stats}).}
\label{tab:family}
\small
\begin{tabular}{lcc}
\toprule
Attention head & 60M & 150M \\
\midrule
Standard & $24.20 \pm 0.13$ & $21.61 \pm 0.03$ \\
Affine coupling, one-way & $24.04 \pm 0.02$ & $21.53 \pm 0.05$ \\
Affine coupling, alternating & $\mathbf{23.92 \pm 0.04}$ & $\mathbf{21.45 \pm 0.06}$ \\
\bottomrule
\end{tabular}
\end{table}

\paragraph{The standalone gain narrows with scale.}
Table~\ref{tab:scaling} compares the standalone alternating coupling against standard attention across scale.
The gain is significant at 60M ($-1.18\%$, $p{=}0.025$, $d{=}3.1$) and 150M ($-0.74\%$, $p{=}0.015$, $d{=}3.4$), but it shrinks as the model grows and is not significant at 455M ($-0.36\%$, three seeds, $p{=}0.19$).
As a standalone change, coupling carries no measurable advantage at 455M.
One caveat qualifies this scale reading: we hold the token budget fixed across sizes (30K steps, about $0.49$B tokens), so the tokens-per-parameter ratio falls from about $8$ at 60M to about $3$ at 150M to about $1$ at 455M, and the larger models are progressively more undertrained. The narrowing of the standalone gain is therefore consistent with both a genuine scaling effect and an increasing training deficit, which this fixed-budget design cannot separate (\S\ref{sec:conclusion}).
A training-budget analysis in the supplementary material sharpens this caveat: tracking the best-so-far validation gap at every evaluation point shows the composition's 455M advantage established early and then stable to slightly widening through the end of training, while the standalone gap closes late, so an optimization-speed reading fits the fading standalone effect but not the composition.

\begin{table}[t]
\small
\centering
\caption{Standalone coupling against standard attention across scale, three seeds, mean$\pm$std. The advantage narrows and is not significant at 455M. Relative reductions are computed from per-seed perplexities, so they can differ slightly from the ratio of the rounded means shown here.}
\label{tab:scaling}
\begin{tabular}{lccc}
\toprule
 & 60M & 150M & 455M \\
\midrule
Standard & $24.20 \pm 0.13$ & $21.61 \pm 0.03$ & $19.56 \pm 0.10$ \\
Alternating coupling & $23.92 \pm 0.04$ & $21.45 \pm 0.06$ & $19.48 \pm 0.07$ \\
\midrule
Relative & $-1.18\%$ & $-0.74\%$ & $-0.36\%$ \\
one-sided $p$ & $0.025$ & $0.015$ & $0.19$ \\
\bottomrule
\end{tabular}
\end{table}

\paragraph{Coupling composes with scoring-stage and projection-level methods.}
This is the central result.
Table~\ref{tab:compose} adds the coupling on top of Differential Attention, Multi-Token Attention, and query-key normalization.
On top of Differential Attention, coupling improves over Differential Attention alone at 150M ($20.92$ against $21.10$, three seeds, $p{=}0.024$, $d{=}3.1$) and at 455M ($18.79$ against $18.89$, six seeds, $p{=}0.003$, $d{=}2.1$).
The standardized effect is $d{=}2.1$ across six seeds; this magnitude indicates direction and robustness rather than a precise estimate, and the comparison survives a Bonferroni correction across our full family of tests (Appendix~\ref{app:stats}).
Held-out test-set perplexity reproduces the comparison ($20.88$ against $21.02$ over the same six seeds, $p{=}0.002$; full test-split table in the supplementary material).
On top of Multi-Token Attention the direction is the same at 150M ($20.85$ against $20.96$) but the effect is not significant ($p{=}0.128$): one seed shows no gain, and we do not claim more than a consistent direction.
Coupling also improves query-key normalization at 60M and 150M (three seeds; $p{=}0.001$ at 150M and $p{=}0.010$ at 60M; the 150M comparison survives Bonferroni correction, the 60M does not); query-key normalization underperforms standard attention in our setup, so this comparison shows the additive direction on a third, independent mechanism. That query-key normalization, and Elliptical Attention which we omit (\S\ref{sec:method}), do not beat standard attention here reflects our short fixed-budget schedule rather than a failure of those methods, so we read the query-key normalization composition as secondary evidence.
Because one small pre-scoring map improves two scoring-stage methods and one projection-level method, and because each composed variant is identical to its base method at initialization, the effect is not explained by added capacity or by a single interacting method.
The compositional gain with Differential Attention is significant at 455M even though the standalone gain has faded there.
At 455M we also evaluate the Multi-Token Attention and query-key normalization compositions: each still improves its base, by $0.20\%$ and $0.48\%$, but neither is significant at that scale ($p{=}0.16$ and $p{=}0.08$, three seeds; query-key normalization is additionally evaluated at 60M, Appendix~\ref{app:stats}), so the additive direction persists while the breadth narrows, and Differential Attention is the only composition that stays significant at 455M.

\begin{table}[t]
\small
\centering
\caption{Composition of the coupling with each base method at 150M and 455M, mean$\pm$std perplexity with one-sided Welch $p$; three seeds, except the 455M Differential Attention composition, which uses six. Each row pairs a method used alone (Base) against that method with the coupling added; bold marks a significant improvement ($p<0.05$). The Differential Attention composition is significant at both scales and survives the Bonferroni correction; at 455M the Multi-Token Attention and query-key normalization compositions improve their base but are not significant, so the additive direction holds while the breadth narrows at scale. Full per-comparison statistics and the Bonferroni analysis are in Appendix~\ref{app:stats}.}
\label{tab:compose}
\begin{tabular}{llccc}

\toprule
Method & Scale & Base & ${+}$ coupling & $p$ \\
\midrule
Differential Attention & 150M & $21.10 \pm 0.08$ & $\mathbf{20.92 \pm 0.02}$ & $0.024$ \\
Differential Attention & 455M & $18.89 \pm 0.04$ & $\mathbf{18.79 \pm 0.05}$ & $0.003$ \\
\midrule
Multi-Token Attention & 150M & $20.96 \pm 0.05$ & $20.85 \pm 0.12$ & $0.128$ \\
Multi-Token Attention & 455M & $19.24 \pm 0.05$ & $19.20 \pm 0.03$ & $0.16$ \\
\midrule
QK-Norm & 150M & $22.25 \pm 0.05$ & $\mathbf{21.96 \pm 0.05}$ & $0.001$ \\
QK-Norm & 455M & $19.82 \pm 0.02$ & $19.73 \pm 0.08$ & $0.08$ \\
\bottomrule
\end{tabular}
\end{table}

\paragraph{The composition holds across training sequence lengths under rotary embeddings.}
To test whether the compositional gain is tied to the 512-token, learned-position setting and how it varies with training length, we train the 455M Differential Attention comparison at sequence lengths 512, 1024, and 2048, all with rotary position embeddings and with the architecture, data, optimizer, step count, and effective batch size unchanged (three seeds each; Figure~\ref{fig:seqlen}).
The composition improves on Differential Attention alone at every length, with every composition seed below every Differential Attention seed: the relative reduction is $1.05\%$ at 512 (one-sided Welch $p{=}0.0035$), $1.54\%$ at 1024 ($p{=}0.0015$), and $2.06\%$ at 2048 ($p{=}0.016$).
The reduction is larger at the longer lengths, but we read this cautiously: it rests on three seeds per length with the highest variance at 2048, and part of the increase in relative terms follows the rising baseline perplexity (the absolute gaps are $0.20$, $0.30$, and $0.42$), so we report a consistent advantage across lengths rather than a tested trend, and the sweep separates neither length from encoding relative to the learned-position headline nor a coupling effect from faster degradation of Differential Attention at longer lengths.
All three rotary lengths are included in our Bonferroni family (Appendix~\ref{app:stats}); at the corrected threshold the 1024 comparison survives while the 512 and 2048 comparisons do not, so we treat the latter two as descriptive, making no claim about lengths beyond 2048.
The sweep also shows the composition functioning under rotary embeddings at 455M, complementing the standalone 60M check in the supplementary material.

\begin{figure}[t]
\centering
\begin{tikzpicture}
\begin{axis}[
  width=0.564\columnwidth, height=0.36\columnwidth,
  xmode=log, log basis x=2,
  xtick={512,1024,2048}, xticklabels={512,1024,2048},
  xlabel={Training sequence length}, ylabel={Validation perplexity},
  xmin=450, xmax=2320, ymin=18.4, ymax=20.6,
  legend style={at={(0.03,0.97)}, anchor=north west, draw=none, font=\footnotesize, fill=none},
  tick label style={font=\footnotesize}, label style={font=\small},
  ymajorgrids=true, grid style={gray!18},
  error bars/y dir=both, error bars/y explicit,
]
\addplot[color=orange!85!black, mark=triangle*, thick, mark size=2.6pt]
  coordinates {(512,18.853)+-(0,0.002) (1024,19.495)+-(0,0.030) (2048,20.251)+-(0,0.170)};
\addlegendentry{Differential Attention}
\addplot[color=blue!70!black, mark=*, thick, mark size=2.2pt]
  coordinates {(512,18.656)+-(0,0.030) (1024,19.196)+-(0,0.053) (2048,19.833)+-(0,0.139)};
\addlegendentry{${+}$ coupling (composition)}
\addplot[only marks, color=orange!85!black, mark=triangle, mark size=2.6pt, forget plot]
  coordinates {(475,18.89)+-(0,0.04)};
\addplot[only marks, color=blue!70!black, mark=o, mark size=2.4pt, forget plot]
  coordinates {(475,18.79)+-(0,0.05)};
\end{axis}
\end{tikzpicture}
\caption{Validation perplexity at 455M for Differential Attention and its composition with the coupling, at rotary-embedding training lengths 512, 1024, and 2048 (filled markers, three seeds; error bars are one standard deviation). The composition is lower at every length, with every composition seed below every Differential Attention seed; the relative reductions are $1.05\%$ ($p{=}0.0035$), $1.54\%$ ($p{=}0.0015$), and $2.06\%$ ($p{=}0.016$). Open markers, offset left for visibility, are the separate learned-positional 512 configuration of the main results (six seeds; Table~\ref{tab:stats}), not a rotary point. Of the three rotary lengths, the 1024 comparison survives the Bonferroni correction (Appendix~\ref{app:stats}); the 512 and 2048 comparisons are significant but do not survive it and are descriptive. Perplexities use the best validation checkpoint, selected symmetrically for both models; at 2048 both overfit in late training (Appendix~\ref{app:stats}).}
\label{fig:seqlen}
\end{figure}

\paragraph{Capacity is not the explanation, and invertibility is what makes coupling reliable.}
To test whether the standalone gain reflects added capacity alone, we compare the alternating coupling against pre-scoring controls that add parameters without its invertible, joint structure (Table~\ref{tab:capacity}, 60M, three seeds, same protocol).
Independent nonlinear projections of $Q$ and $K$, which add nonlinear capacity without coupling them, do not improve on standard attention ($24.13$ against $24.20$, $p{=}0.22$).
A non-invertible cross MLP that mixes the two reaches $24.04$ but carries more parameters than the coupling ($+0.43\%$ against $+0.33\%$), and a linear mixing variant with five times the coupling's parameter budget ($+1.74\%$) is the worst of the set ($24.34$).
The alternating coupling ($23.92$) is significantly better than every one of these controls, including the higher-parameter cross MLP ($p{=}0.035$) and the highest-parameter linear mix ($p{<}0.001$).
The gain tracks the invertible, joint structure rather than the parameter count: the variants with the most parameters are not the best, and because each composed variant is identical to its base method at initialization, the compositional gains are not a starting-point artifact either.
At 455M, the scale of the central result, two further controls on top of Differential Attention complete this attribution (three seeds each, identical protocol; supplementary material).
The non-invertible cross MLP reaches $18.89 \pm 0.11$ against $18.89 \pm 0.04$ for Differential Attention alone ($p{=}0.51$), reproducing none of the composition gain, so added pre-scoring capacity does not explain it.
A simultaneous variant of the coupling itself, identical in parameterization but with the invertibility-guaranteeing sequential update replaced by a simultaneous one, retains the 455M gain ($18.78 \pm 0.07$) yet is dramatically worse than Differential Attention alone at 150M ($22.17 \pm 0.04$ against $21.10 \pm 0.08$, $p{=}0.0001$), below even standard attention, despite smooth and stable training.
The gain therefore tracks the joint pre-scoring coupling rather than parameter count or invertibility alone, and the invertible sequential form is the only pre-scoring variant that improves its base method at every scale we test: invertibility is the property that makes the coupling reliable across scales.

\begin{table}[t]
\small
\centering
\caption{Capacity controls at 60M, three seeds, mean$\pm$std perplexity, with parameter overhead over standard attention. Adding parameters without the invertible, joint coupling does not reproduce the gain, and the variants with the most parameters are not the best. The alternating coupling is significantly better than every control (one-sided Welch, $p<0.05$). Matched controls at 455M, including a non-invertible simultaneous variant of the coupling itself, are in the supplementary material.}
\label{tab:capacity}
\begin{tabular}{lcc}
\toprule
Pre-scoring variant & Params & 60M perplexity \\
\midrule
Standard & $+0.00\%$ & $24.20 \pm 0.13$ \\
Independent nonlinear $Q,K$ & $+0.11\%$ & $24.13 \pm 0.03$ \\
Non-invertible cross MLP & $+0.43\%$ & $24.04 \pm 0.07$ \\
Linear mix & $+1.74\%$ & $24.34 \pm 0.05$ \\
Alternating coupling & $+0.33\%$ & $\mathbf{23.92 \pm 0.04}$ \\
\bottomrule
\end{tabular}
\end{table}

\paragraph{Generalization to a second corpus.}
To test whether the effect is specific to WikiText-103, we repeat the 60M comparison on a FineWeb-Edu slice (Table~\ref{tab:fineweb}).
The standalone coupling lowers perplexity from $45.30$ to $44.78$ (two seeds, both favorable), and the composition with Differential Attention reaches $44.35$, below Differential Attention alone ($45.12$) and standard attention ($45.30$).
With two seeds and no significance test this second-corpus comparison is descriptive only.

\begin{table}[t]
\small
\centering
\caption{FineWeb-Edu cross-corpus check at 60M. Per-seed and mean validation perplexity over two seeds; lower is better. Both seeds favor coupling in every row. With two seeds we report this as descriptive, with no significance test.}
\label{tab:fineweb}
\begin{tabular}{lccc}
\toprule
Configuration & seed 1 & seed 2 & mean \\
\midrule
Standard & $45.34$ & $45.25$ & $45.30$ \\
Alternating coupling & $44.66$ & $44.89$ & $\mathbf{44.78}$ \\
\midrule
Differential Attention & $45.31$ & $44.94$ & $45.12$ \\
\quad ${+}$ coupling & $44.20$ & $44.50$ & $\mathbf{44.35}$ \\
\bottomrule
\end{tabular}
\end{table}

\paragraph{Zero-shot transfer.}
We evaluate the 150M and 455M checkpoints zero-shot on LAMBADA last-token accuracy and HellaSwag length-normalized accuracy (Table~\ref{tab:downstream}).
At 455M we evaluate every seed (six per side for the Differential pair, three for standard attention and the standalone coupling): the composition is best on both tasks, leading Differential Attention on LAMBADA ($0.272$ against $0.263$, one-sided Welch $p{=}0.04$) and on HellaSwag ($0.310$ against $0.306$, $p{=}0.15$), and leading standard attention on both ($p{=}0.007$ and $p{=}0.04$).
The standalone coupling falls below standard attention on both tasks at 455M, consistent with its faded perplexity gain; under multi-seed evaluation, Differential Attention remains above standard attention on both tasks.
At 150M the evaluations remain single-seed (seed 42), so we report no significance there; the composition leads LAMBADA and the standalone coupling is highest on HellaSwag.
These transfer accuracies are supporting evidence outside the Bonferroni-corrected perplexity family: a small perplexity gap on WikiText-103 need not align with zero-shot accuracy on a different distribution.

\begin{table}[t]
\small
\centering
\caption{Zero-shot transfer at 150M (single seed, seed~42) and 455M (mean$\pm$std; six seeds per side for the Differential pair, three for standard and the standalone coupling); 2{,}000 examples per task. LAMBADA last-token accuracy and HellaSwag length-normalized accuracy; higher is better. A coupling variant is best in every column (bold). At 455M the composition is best on both tasks, and the standalone coupling falls below standard, consistent with its faded perplexity gain.}
\label{tab:downstream}
\begin{tabular}{lcccc}
\toprule
 & \multicolumn{2}{c}{LAMBADA} & \multicolumn{2}{c}{HellaSwag} \\
\cmidrule(lr){2-3}\cmidrule(lr){4-5}
Configuration & 150M & 455M & 150M & 455M \\
\midrule
Standard & $0.259$ & $0.254 \pm 0.006$ & $0.313$ & $0.303 \pm 0.004$ \\
Alternating coupling & $0.269$ & $0.252 \pm 0.003$ & $\mathbf{0.317}$ & $0.297 \pm 0.006$ \\
Differential Attention & $0.274$ & $0.263 \pm 0.007$ & $0.307$ & $0.306 \pm 0.007$ \\
\quad ${+}$ coupling & $\mathbf{0.278}$ & $\mathbf{0.272 \pm 0.010}$ & $0.315$ & $\mathbf{0.310 \pm 0.006}$ \\
\bottomrule
\end{tabular}
\end{table}

\paragraph{Where coupling helps, and where it does not.}
We use multi-query associative recall~\citep{arora2024zoology} to characterize the mechanism, with a small model ($d{=}256$, $H{=}4$, $L{=}6$), sweeping the number of key-value pairs against sequence length over three seeds (Table~\ref{tab:mqar}).
At two key-value pairs every variant saturates.
At four pairs the standalone coupling is both higher and far more stable across seeds than standard attention ($0.99{\pm}0.01$ against $0.92{\pm}0.14$ at length 256), consistent with the reading that it forms a retrieval head from a broader, lower-variance prior.
At eight pairs, the hardest retrieval load, neither standalone head is reliable: standard attention is high-variance (means $0.55$ to $0.79$ with seed spreads up to $0.40$), and the coupling does not help on average and collapses at the densest setting, eight pairs at length 64, to $0.18{\pm}0.00$ across all three seeds, which we read as its smoothing suppressing the single sharp pointer this regime needs.
The composition resolves this: adding Differential Attention to either head restores recall to $1.00$ across the entire grid, so the scoring-stage denoising supplies exactly the sharp selection that the standalone coupling smooths away, and the coupling adds no retrieval cost once that selection is in place.
This is a single small synthetic setup and the mechanism in full language modeling may differ; we therefore base the compositional recommendation on the scale result, with which this characterization is consistent.

\begin{table}[t]
\small
\centering
\caption{Multi-query associative-recall accuracy, mean$\pm$std over three seeds, on a small model ($d{=}256$, $H{=}4$, $L{=}6$); higher is better. All variants reach $1.00$ at two key-value pairs (row omitted). At four pairs the standalone coupling is higher and far more stable across seeds than standard attention; at eight pairs neither standalone head is reliable, and the coupling collapses at the densest cell. Composing either standalone head with Differential Attention restores recall to $1.00$ on every cell (all $\ge 0.999$).}
\label{tab:mqar}
\begin{tabular}{llcccc}
\toprule
KV pairs & Seq.\ len & Standard & Coupling & Diff.\ Attn & ${+}$Coupling \\
\midrule
4 & 64  & $0.98 \pm 0.02$ & $1.00 \pm 0.00$ & $1.00$ & $1.00$ \\
4 & 128 & $0.94 \pm 0.05$ & $1.00 \pm 0.00$ & $1.00$ & $1.00$ \\
4 & 256 & $0.92 \pm 0.14$ & $0.99 \pm 0.01$ & $1.00$ & $1.00$ \\
\midrule
8 & 64  & $0.55 \pm 0.40$ & $0.18 \pm 0.00$ & $1.00$ & $1.00$ \\
8 & 128 & $0.79 \pm 0.10$ & $0.72 \pm 0.37$ & $1.00$ & $1.00$ \\
8 & 256 & $0.73 \pm 0.34$ & $0.51 \pm 0.41$ & $1.00$ & $1.00$ \\
\bottomrule
\end{tabular}
\end{table}

\paragraph{Attention rank at depth.}
We probe the 24-layer 455M models on validation sequences (full protocol in the supplementary material).
The effective rank of the pre-softmax logit matrix is essentially identical across all variants (about $14.4$) and is set by the causal mask, so the coupling does not change it.
The effective rank of the post-softmax attention map, averaged over the late layers (L16 to L23), is modestly higher for the standalone coupling than for standard attention ($53.5$ against $46.2$), while the Differential Attention pair differs little ($45.3$ against $42.5$).
We report this as a modest effect that is consistent with the small standalone gain, and we do not claim that coupling escapes a rank-collapse bound.

\paragraph{Practical guidance.}
The reading we take to practice is to use coupling compositionally rather than on its own.
At the scales we test, a practitioner who already uses a scoring-stage method such as Differential Attention can add the alternating coupling for a small parameter overhead ($0.33\%$ at 60M, $0.13\%$ at 455M) and obtain a small but consistent further reduction in perplexity that is significant at 455M, at a measured cost that shrinks at the scale where the result lives: about $8$ to $10\%$ training throughput over Differential Attention alone at 60M and sequence length 512, falling to $7.4\%$ at 455M and to $4.1\%$ at 455M with length 2048, with peak-memory overheads of about $17\%$, $23\%$, and $11\%$ in the three settings, benchmarked on an H100 (supplementary material); whether that trade is worthwhile depends on the deployment, and a fused coupling kernel would narrow the latency gap toward the parameter overhead.
Using coupling as a standalone replacement for standard attention is worthwhile mainly below the 150M range, where its standalone gain is significant; at 455M the standalone gain is not, so we do not recommend it there.

%% file: sections/related.tex
\section{Related Work}
\label{sec:related}

\paragraph{Pre-scoring transformations of queries and keys.}
Several methods act on $Q$ and $K$ before the dot product.
Query-key normalization rescales each vector to stabilize logits at scale~\citep{henry2020query}, and the normalized Transformer places all representations on a hypersphere so that every logit is a cosine similarity~\citep{loshchilov2024ngpt}.
These operations are per-vector and norm-only; they do not let the query and the key inform each other. Query-key normalization is moreover head-local, whereas the normalized Transformer's hypersphere constraint is global, which makes the latter a less separable target for a per-head coupling.
Talking-Heads attention applies learned linear maps across heads both before and after the softmax~\citep{shazeer2020talking}, which mixes heads at the scoring stage rather than coupling the two vectors within a head.
A separate family replaces the dot product itself: the Synthesizer uses learned or random score maps~\citep{tay2021synthesizer}, and the Performer approximates the softmax kernel with random features~\citep{choromanski2021rethinking}; both discard the standard dot product that coupling leaves intact, so they are not pre-scoring transformations in our sense.
Closest to a learned joint comparison, Elliptical Attention replaces the inner product with a learned positive-definite quadratic form, a shared bilinear metric between queries and keys~\citep{elliptical2024}; this is a global, input-independent learned metric on the $(q,k)$ pair, whereas coupling is a nonlinear, alternating, invertible map that reshapes the joint representation in an input-adaptive way rather than reweighting directions uniformly.
Our coupling differs from all of these: it is a learned, joint, invertible map of $Q$ and $K$ within each head that keeps the exact softmax dot product, and it is the identity at initialization so it can only add to a standard head.

\paragraph{Scoring-stage methods that coupling composes with.}
A second family modifies the scoring stage after the product.
Differential Attention subtracts two softmax maps to cancel common-mode noise~\citep{ye2025diff}, Multi-Token Attention convolves over the score map to share information across nearby positions~\citep{golovneva2025multitoken}, and gated attention reweights the attention output with a learned gate~\citep{gated2025attention}.
These methods are complementary to coupling rather than alternatives to it, and our central experiments place coupling on top of the first two.

\paragraph{Concurrent work on a nonlinear query-key interaction.}
Recent work also questions the linearity of the query-key map.
One line replaces the linear query, key, or value projections with nonlinear ones~\citep{nexusformer2026}, and another adds a nonlinear residual to the query alone~\citep{beyondlinearity2026}.
M\"{o}biusAttention introduces a nonlinear M\"{o}bius transform into the score computation~\citep{mobius2025attention}; because it acts on the logit rather than on the projections, it is a scoring-stage method that coupling composes with in principle, not a pre-scoring transformation.
Coupling is distinct on three counts that matter for our claims: it transforms $Q$ and $K$ jointly rather than independently, it is invertible by construction, and we show it composes additively with scoring-stage methods, a property these works do not study.

\paragraph{Coupling flows and dynamical views of attention.}
The alternating affine coupling is the construction used in real-valued non-volume-preserving flows~\citep{dinh2017density}, and coupling flows of this kind are universal approximators of invertible smooth maps given sufficient depth~\citep{teshima2020coupling}; we use this construction for its expressivity, not as a density model.
A separate line views attention through dynamical systems and energy minimization, including modern Hopfield networks~\citep{ramsauer2021hopfield} and analyses of attention as interacting particles~\citep{geshkovski2024mathematical}.
This view motivated our starting point, an additive symplectic update; because a volume-preserving map of this kind is provably restricted in the family it can represent, we adopt the non-volume-preserving affine coupling, and the gain comes from that coupling rather than from any conservation law.

\paragraph{Attention pathologies.}
Pure self-attention loses rank with depth~\citep{dong2021attention}, and the attention matrix exhibits a widening spectral gap that is specific to the softmax~\citep{nait2024spectral}.
We measure attention rank directly at depth and report a modest effect rather than a resolution of collapse (\S\ref{sec:experiments}); we do not rest the method on this analysis.

%% file: sections/conclusion.tex
\section{Conclusion}
\label{sec:conclusion}

We studied coupled query-key dynamics, a pre-scoring transformation that evolves queries and keys jointly through a shared invertible coupling before standard scoring, realized as an alternating affine coupling that is the identity at initialization.
A controlled ablation isolates the alternating, non-volume-preserving form as the one that helps and separates it from added capacity; the additive, volume-preserving update we began with is provably more restricted, which is why we adopt the affine coupling.
As a standalone change, coupling lowers perplexity at 60M and 150M with margins that pass significance testing ($p<0.05$) but loses that margin by 455M.
The behavior that holds at scale is compositional: placed on top of Differential Attention, coupling improves on it at both 150M and 455M, and the same additive direction appears on top of Multi-Token Attention (directional) and query-key normalization (significant at 150M).
We therefore present coupling not as a standalone competitor but as a composable pre-scoring transformation that adds to scoring-stage and projection-level methods; we use the word axis only in the empirical sense that the addition is consistent, not as a claim of structural independence.

\paragraph{Limitations.}
The standalone gain fades with scale, and our largest model is 455M; whether the compositional gain persists beyond this scale is open.
Our scale sweep holds the token budget fixed at about $0.49$B tokens, so the tokens-per-parameter ratio falls from about $8$ at 60M to about $1$ at 455M, with the largest model well below the compute-optimal ratio; the narrowing of the standalone gain is consistent with both a scaling effect and increasing undertraining, which our fixed-budget design cannot separate.
The composition with Multi-Token Attention is directional rather than significant at both 150M and 455M.
The compositional result is established for one scoring-stage method, Differential Attention, at the scale where the standalone gain has faded; query-key normalization is not a competitive baseline in our setup, so whether coupling is universally additive on scoring-stage methods or specific to Differential Attention is open, and the FineWeb-Edu check uses two seeds with no significance claim.
On associative recall the standalone coupling does not improve the hardest retrieval load and collapses at the densest setting, since it smooths the sharp single-pointer head that the task needs; composing it with Differential Attention restores full recall there, consistent with the compositional reading.
Most composition experiments use learned positional embeddings at sequence length 512; a rotary-embedding sweep at lengths 512, 1024, and 2048 shows the Differential Attention composition improving at every length, with a larger margin at the longer lengths that we report as a consistent pattern rather than a tested trend (\S\ref{sec:experiments}). That sweep uses three seeds per length and does not separate length from encoding relative to the learned-position headline, and we have not tested lengths beyond 2048.
Our experiments use one model family, and the rank analysis at depth shows only a modest effect rather than a resolution of rank collapse.

\paragraph{Future work.}
The natural next steps are to push the compositional comparison to larger scale, to tune each scoring-stage method and the coupling jointly, and to test the coupling across additional model families.
More broadly, the results suggest that a small, invertible, identity-initialized transformation of queries and keys before scoring can add to methods that act after scoring, and that reporting where such a mechanism fails is as informative as reporting where it helps.

%% file: sections/appendix.tex
\section{Experimental Setup}
\label{app:setup}

\paragraph{Model configurations.}

\begin{table}[h]
\small
\centering
\caption{Model configurations used in experiments.}
\label{tab:configs}
\begin{tabular}{lccccc}
\toprule
\textbf{Config} & $d_\text{model}$ & $n_\text{heads}$ & $n_\text{layers}$ & $d_\text{ff}$ & \textbf{Params} \\
\midrule
Tiny   & 256  & 4  & 6  & 1024 & $\sim$6M$^\dagger$ \\
Small  & 512  & 8  & 8  & 2048 & $\sim$60M \\
Medium & 768  & 12 & 12 & 3072 & $\sim$153M \\
Large  & 1024 & 16 & 24 & 4096 & $\sim$455M \\
\bottomrule
\end{tabular}
\end{table}

All models share pre-norm RMSNorm, SwiGLU feed-forward layers, tied input/output embeddings, and learned positional embeddings.\\
$^\dagger$Parameter counts depend on vocabulary size: Tiny is used only for MQAR (vocab=64); the other counts are for WikiText-103 (vocab=50{,}257).

At the fixed WikiText-103 budget of 30K steps (about $0.49$B tokens), the tokens-per-parameter ratio is about $8.2$ (Small, 60M), $3.2$ (Medium, 150M), and $1.1$ (Large, 455M); the largest model is well below the compute-optimal ratio, so larger models in our sweep are progressively more undertrained, a confound flagged in the Limitations (\S\ref{sec:conclusion}).

\paragraph{Training hyperparameters.}
\label{app:hyperparams}

\emph{MQAR experiments.}
AdamW ($\beta_1{=}0.9$, $\beta_2{=}0.95$), $\text{lr}{=}3{\times}10^{-4}$, weight decay 0.01, gradient clipping 1.0, cosine schedule, 500-step warmup, 10K--20K total steps, batch size 64.

\emph{WikiText-103 (small, 60M).}
AdamW ($\beta_1{=}0.9$, $\beta_2{=}0.95$), $\text{lr}{=}6{\times}10^{-4}$, weight decay 0.1, gradient clipping 1.0, cosine schedule with 2K-step warmup, 30K total steps, effective batch size 32 (micro-batch 8, gradient accumulation 4), mixed precision (fp16), sequence length 512.

\emph{WikiText-103 (medium, 150M).}
Same as small except: $\text{lr}{=}3{\times}10^{-4}$, 30K total steps, gradient checkpointing enabled, effective batch size 32 (micro-batch 8, gradient accumulation 4; coupling variant: micro-batch 4, gradient accumulation 8).

\emph{WikiText-103 (large, 455M).}
Same as medium except: $\text{lr}{=}2{\times}10^{-4}$, effective batch size 32 (micro-batch 4, gradient accumulation 8; coupling variant: micro-batch 2, gradient accumulation 16).
The rotary length sweep (\S\ref{sec:experiments}) uses sequence lengths 512, 1024, and 2048 with rotary position embeddings, three seeds each (42, 137, 2024), the same optimizer and step count, and effective batch 32; at 2048 the micro-batch is halved and accumulation doubled to preserve the effective batch within memory.

\paragraph{Variant-specific hyperparameters.}
Table~\ref{tab:variant_hparams} lists the hyperparameters specific to each attention variant, including the complete specification of the coupling; the coupling introduces no other hyperparameters and leaves those of the base methods unchanged.

\begin{table}[h]
\small
\centering
\caption{Variant-specific hyperparameters. The coupling rows fully specify the alternating affine coupling of Eqs.~\eqref{eq:coupleK}--\eqref{eq:coupleQ}.}
\label{tab:variant_hparams}
\begin{tabular}{llc}
\toprule
\textbf{Variant} & \textbf{Hyperparameter} & \textbf{Value} \\
\midrule
\multirow{5}{*}{Alternating coupling}
  & Coupling rounds & 1 (Eqs.~\eqref{eq:coupleK}--\eqref{eq:coupleQ}) \\
  & Conditioning nets $s,t$ & two-layer, SiLU, output width $2 d_k$ \\
  & Net placement & one $(s,t)$ pair per update per layer, per head \\
  & Output-layer init & zero (exact identity at initialization) \\
  & Log-scale clamp & $[-3, 3]$ \\
\midrule
\multirow{2}{*}{GQA}
  & KV head ratio & $1/4$ of query heads \\
  & Group size & 4 queries per KV head \\
\midrule
\multirow{2}{*}{Differential Attention}
  & Sub-head dimension $d_s$ & $d_k / 2$ \\
  & $\lambda$ init & $0.8 - 0.6 \exp(-0.3 \ell)$ (layer $\ell$) \\
\bottomrule
\end{tabular}
\end{table}

\paragraph{Software and hardware.}
All reported results were produced on NVIDIA H100 GPUs with PyTorch 2.10 and mixed-precision (fp16) training.
All attention variants are drop-in replacements behind a shared interface, so every comparison uses identical training infrastructure and differs only in the attention head. All models use learned positional embeddings by default; the RoPE ablation (supplementary material) and the 455M longer-context comparison (\S\ref{sec:experiments}) replace these with Rotary Position Embeddings~\citep{su2024roformer}.

\section{Per-Seed Statistical Tests}
\label{app:stats}

All perplexity values in Table~\ref{tab:stats} are the exponential of the best validation loss; significance uses a one-sided Welch's $t$-test with the alternative that the treatment perplexity is lower, and the $n$ column gives seeds per comparison.

\begin{table}[h]
\centering
\caption{Per-comparison statistics for all main claims. $\Delta = $ treatment PPL $-$ baseline PPL
  (negative $=$ improvement). Significance: $^{**}p < 0.01$, $^{*}p < 0.05$, $^{\dagger}p < 0.10$.}
\label{tab:stats}
\small
\resizebox{\columnwidth}{!}{%
\begin{tabular}{llrrrrrll}
\toprule
\textbf{Comparison} & \textbf{Scale} & $n$ &
  \textbf{Baseline mean$\pm$std} & \textbf{Treatment mean$\pm$std} &
  $\Delta$ & $p$ & $d$ & \textbf{Note} \\
\midrule
\multirow{3}{*}{\parbox{3cm}{1.\ Standalone coupling\\(vs.\ standard)}}
  & 60M  & 3 & $24.20 \pm 0.13$ & $23.92 \pm 0.04$ & $-0.28$ & $0.025^{*}$  & 3.09 & \\
  & 150M & 3 & $21.61 \pm 0.03$ & $21.45 \pm 0.06$ & $-0.16$ & $0.015^{*}$  & 3.39 & \\
  & 455M & 3 & $19.56 \pm 0.10$ & $19.48 \pm 0.07$ & $-0.07$ & $0.19$ & 0.83 & not significant \\
\midrule
\multirow{3}{*}{\parbox{3cm}{2.\ Composition:\\Differential Attention}}
  & 150M & 3 & $21.10 \pm 0.08$ & $20.92 \pm 0.02$ & $-0.17$ & $0.024^{*}$  & 3.12 & \\
  & 455M & 6 & $18.89 \pm 0.04$ & $18.79 \pm 0.05$ & $-0.10$ & $0.003^{**}$ & 2.09 & \\
  & 455M & 3 & $19.50 \pm 0.03$ & $19.20 \pm 0.05$ & $-0.30$ & $0.0015^{**}$ & 6.95 & seq 1024, RoPE \\
\midrule
\multirow{2}{*}{\parbox{3cm}{3.\ Composition:\\Multi-Token Attn.}}
  & 150M & 3 & $20.96 \pm 0.05$ & $20.85 \pm 0.12$ & $-0.11$ & $0.128$      & 1.18 & not significant \\
  & 455M & 3 & $19.24 \pm 0.05$ & $19.20 \pm 0.03$ & $-0.04$ & $0.16$       & 0.92 & not significant \\
\midrule
\multirow{3}{*}{\parbox{3cm}{4.\ Composition:\\QK-Norm}}
  & 60M  & 3 & $24.24 \pm 0.00$ & $24.12 \pm 0.03$ & $-0.13$ & $0.010^{**}$ & 5.53 & \\
  & 150M & 3 & $22.25 \pm 0.05$ & $21.96 \pm 0.05$ & $-0.29$ & $0.001^{**}$ & 5.53 & \\
  & 455M & 3 & $19.82 \pm 0.02$ & $19.73 \pm 0.08$ & $-0.09$ & $0.08^{\dagger}$ & 1.68 & not significant \\
\midrule
\multirow{4}{*}{\parbox{3cm}{5.\ Ablation ladder}}
  & 60M  (std$\to$coup)       & 3 & $24.20 \pm 0.13$ & $24.04 \pm 0.02$ & $-0.16$ & $0.077^{\dagger}$ & 1.79 & marginal \\
  & 60M  (coup$\to$alt)       & 3 & $24.04 \pm 0.02$ & $23.92 \pm 0.04$ & $-0.12$ & $0.006^{**}$      & 4.35 & \\
  & 150M (std$\to$coup)       & 3 & $21.61 \pm 0.03$ & $21.53 \pm 0.05$ & $-0.08$ & $0.048^{*}$       & 1.92 & \\
  & 150M (coup$\to$alt)       & 3 & $21.53 \pm 0.05$ & $21.45 \pm 0.06$ & $-0.09$ & $0.068^{\dagger}$ & 1.54 & marginal \\
\bottomrule
\end{tabular}
}
\end{table}

\noindent We use one-sided tests because the directional hypothesis that coupling lowers perplexity was established from the 60M standalone results before the larger runs were launched, and from the composition direction before the composition runs; we report the non-significant outcomes without changing the test direction post hoc. The two-sided equivalents double the reported $p$-values; the Bonferroni analysis below uses the one-sided values. Even two-sided, the headline 455M Differential composition remains significant ($p\approx0.005$).
All $p$-values and effect sizes are computed from the unrounded per-seed perplexities; for the headline 455M Differential comparison the six per-seed perplexities are $18.9054$, $18.9325$, $18.9110$, $18.8679$, $18.8176$, $18.8812$ for Differential Attention and $18.7864$, $18.8188$, $18.7673$, $18.7224$, $18.7666$, $18.8728$ for the composition, which reproduce one-sided $p{=}0.0026$ (rounded to $0.003$ in Table~\ref{tab:stats}) and $d{=}2.1$. Per-seed perplexities for the remaining comparisons (the two 455M breadth pairs and the three rotary lengths) are in the supplementary material; for three-seed rows the Cohen's $d$ magnitudes indicate direction rather than stable estimates. At length 2048 both models overfit late in training, so that comparison uses each model's best validation checkpoint, selected independently.
Applying a Bonferroni correction across the seventeen comparisons in our family, the fifteen in this table plus the 512 and 2048 rotary lengths of the length sweep, sets $\alpha{=}0.05/17{\approx}0.0029$. Three comparisons remain significant at that threshold: the 455M Differential composition at length 512 ($p{=}0.0026$), the same composition at length 1024 with rotary embeddings ($p{=}0.0015$), and the 150M query-key normalization composition ($p{=}0.001$). All others do not, including the 512 and 2048 rotary compositions ($p{=}0.0035$ and $p{=}0.016$), the 60M query-key normalization composition ($p{=}0.010$), and the coupling-to-alternating ablation ($p{=}0.006$).

%% file: sections/supp_controls.tex
\section{Invertibility Controls at Scale and Training-Budget Analysis}
\label{app:controls}

This section reports the two controls and the training-budget analysis added in
response to reviewer requests during the ACML 2026 discussion period. The
comparisons here are post-hoc additions and therefore lie outside the
pre-registered Bonferroni family of the main paper; we report them with per-seed
values and treat them as attribution evidence for the composition result rather
than as additional headline claims.

\paragraph{Two matched controls on top of Differential Attention.}
The capacity table in the main paper compares pre-scoring controls at 60M.
Two questions remain open there: whether extra pre-scoring capacity could
reproduce the 455M composition gain, and whether invertibility itself, rather
than the bidirectional affine interaction, is the operative property.
We answer both with controls trained on top of Differential Attention under the
identical protocol (data, steps, optimizer, sequence length 512, three seeds).

The first control is the non-invertible cross-MLP of the capacity table: a
joint two-layer MLP over the stacked $(q,k)$ pair with a learned step size,
zero-initialized so it is the exact identity at initialization, and with
slightly more parameters than the coupling.

The second control is the minimal-difference variant of the coupling itself.
The alternating coupling of the main paper is sequential: the query first
conditions the key update, and the \emph{updated} key then conditions the query
update, which yields the triangular structure that guarantees invertibility.
The simultaneous control keeps the conditioning networks, the parameter count
(equal to the last parameter), the clamping, and the identity-at-init guarantee
of the coupling, and changes only the update order: both vectors are updated
from the pre-update partner,
\begin{align}
k' &= k \odot \exp\!\big(s_\theta(q)\big) + t_\theta(q),\\
q' &= q \odot \exp\!\big(s_\phi(k)\big) + t_\phi(k),
\end{align}
where $s_\phi, t_\phi$ read the original $k$ rather than $k'$.
This removes the invertibility guarantee, since recovering $(q,k)$ from
$(q',k')$ requires solving a coupled nonlinear system with no bijectivity
guarantee, while preserving the bidirectional interaction and the affine
scaling.

\begin{table}[h]
\centering
\caption{Invertibility and capacity controls on top of Differential Attention,
best validation perplexity, mean$\pm$std. One-sided Welch $p$ is against
Differential Attention alone at the same scale (six seeds at 455M, three at
150M); for the two controls that are \emph{worse} than the base, the reported
$p$ is for the reversed one-sided test. All variants are exactly their base
method at initialization.}
\label{tab:controls}
\begin{tabular}{llccc}
\toprule
Scale & Pre-scoring variant on Diff.\ Attn. & Perplexity & $\Delta$ vs.\ base & $p$ \\
\midrule
455M & none (Differential Attention, 6 seeds) & $18.89 \pm 0.04$ & & \\
455M & non-invertible cross-MLP (3 seeds) & $18.89 \pm 0.11$ & $+0.002$ & $0.51$ \\
455M & simultaneous coupling (3 seeds) & $18.78 \pm 0.07$ & $-0.110$ & $0.042$ \\
455M & invertible alternating coupling (6 seeds) & $\mathbf{18.79 \pm 0.05}$ & $-0.097$ & $0.003$ \\
\midrule
150M & none (Differential Attention, 3 seeds) & $21.10 \pm 0.08$ & & \\
150M & simultaneous coupling (3 seeds) & $22.17 \pm 0.04$ & $+1.078$ & $0.0001$ (worse) \\
150M & invertible alternating coupling (3 seeds) & $\mathbf{20.92 \pm 0.02}$ & $-0.175$ & $0.024$ \\
\bottomrule
\end{tabular}
\end{table}

\paragraph{Result: a dissociation.}
Table~\ref{tab:controls} gives the two controls' outcomes.
The cross-MLP reproduces none of the 455M gain: it is statistically
indistinguishable from Differential Attention alone ($p{=}0.51$), so added
pre-scoring capacity does not explain the composition result.
The simultaneous coupling dissociates the two remaining factors.
At 455M it reproduces the gain (better than the base, $p{=}0.042$, and
statistically indistinguishable from the invertible composition), so
invertibility per se is not the sole cause of the gain at that scale; the joint
bidirectional coupling is.
At 150M, however, the same simultaneous variant is dramatically worse than
Differential Attention alone, and worse than standard attention, despite
smooth and stable training at every seed: the failure is a bad optimum, not
divergence.
The invertible sequential form is the only pre-scoring variant that improves
its base method at every scale we test.
We therefore read invertibility as the design property that makes bidirectional
query-key coupling consistently safe and beneficial across scales, rather than
as the isolated cause of the 455M gain, and the main text states the claim in
this form.
Per-seed best validation perplexities: simultaneous coupling
$18.7113/18.7737/18.8439$ at 455M and $22.1307/22.1919/22.1981$ at 150M;
cross-MLP $18.9875/18.7631/18.9126$ at 455M.

\paragraph{Training-budget analysis.}
The models in this study are undertrained relative to compute-optimal token
budgets (about $1.1$ tokens per parameter at 455M), which raises the concern
that the composition gain reflects optimization speed in an undertrained
regime rather than a persistent advantage.
An optimization-speed advantage predicts a gap that closes as training
progresses.
We test this on the existing training runs by computing, at every evaluation
point (every 1{,}000 steps), the best-so-far validation perplexity of each
variant, i.e., the value the paper's best-checkpoint comparison would report if
training had stopped at that budget, and tracking the relative gap between the
composition and Differential Attention alone (three seeds per side).

\begin{table}[h]
\centering
\caption{Best-so-far relative validation-perplexity gap (mean over three
seeds; negative favors the coupling variant) as a function of training budget
at 455M. The composition's advantage is established early and remains stable
to slightly widening through the end of training (late-half trend
$-0.02\%$ per 1{,}000 steps), while the standalone gap closes late
($+0.02\%$ per 1{,}000 steps), matching the faded standalone effect reported
in the main paper.}
\label{tab:trajectory}
\begin{tabular}{lcccc}
\toprule
455M comparison & 8K steps & 15K steps & 22K steps & 29K steps \\
\midrule
Composition vs.\ Differential Attention & $-0.37\%$ & $-0.66\%$ & $-0.65\%$ & $-0.65\%$ \\
Standalone coupling vs.\ standard & $-0.31\%$ & $-0.73\%$ & $-0.79\%$ & $-0.34\%$ \\
\bottomrule
\end{tabular}
\end{table}

Table~\ref{tab:trajectory} shows the opposite of the optimization-speed
prediction for the composition: the relative gap reaches $-0.66\%$ by 15K
steps and stays essentially constant through 29K steps.
The same diagnostic applied to the standalone coupling shows the gap closing
in late training, consistent with the faded standalone effect at 455M, so the
analysis distinguishes a persistent advantage from a transient one rather than
being insensitive to the difference.
This does not substitute for a compute-optimal 455M run, which is beyond our
budget; it tests, and rejects, the specific prediction that the composition
advantage shrinks with more training.

%% file: sections/supp_overhead.tex
\section{Overhead Benchmarks}
\label{app:overhead}

\paragraph{Throughput.}
Each coupling round applies a two-layer per-head MLP of width $d_k$; across all heads this adds on the order of $d\,d_k$ multiply-accumulates per token per layer and does not change the asymptotic $\mathcal{O}(n^2 d_k)$ attention cost.
The parameter overhead is about $0.3\%$ at 60M (Table~\ref{tab:params}).
Table~\ref{tab:throughput} reports the measured cost at 60M on a single H100 (batch 8, forward and backward, tokens per second relative to standard attention).
Because the per-head coupling MLPs are not fused, the realized overhead exceeds the parameter overhead: the standalone coupling runs at about $90\%$ of standard throughput at sequence length 512 and about $85\%$ at length 2048, and the Differential Attention composition runs at about $92\%$ and $90\%$ of Differential Attention alone at the two lengths.
Peak memory rises by about $22\%$ for the standalone coupling and about $17\%$ for the composition over its base, both at length 512.
At length 2048 Differential Attention is itself the costlier component (about $0.68$ relative throughput and $1.76\times$ peak memory against standard), so the coupling is not the dominant cost in the composition.
A fused kernel would narrow the gap toward the parameter overhead, and implementing one is left to future work.

\begin{table}[h]
\centering
\caption{Parameter comparison at 60M scale. The alternating affine coupling variant uses per-head coupling MLPs applied in an alternating pattern, adding approximately 0.3\% parameters over standard attention.}
\label{tab:params}
\begin{tabular}{lrrr}
\toprule
\textbf{Attention} & \textbf{Total Params} & \textbf{Variant-Specific} & \textbf{Overhead} \\
\midrule
Standard                     & 60{,}343{,}296 & 0        & n/a       \\
Affine coupling, one-way     & 60{,}408{,}896 & 65{,}600 & $+$0.11\% \\
Affine coupling, alternating & 60{,}541{,}952 & 198{,}656 & $+$0.33\% \\
GQA                          & 57{,}197{,}568 & 0        & $-$5.21\% \\
Diff Attention               & 60{,}343{,}360 & 64       & $+$0.00\% \\
\bottomrule
\end{tabular}
\end{table}

The one-way row is the intermediate, non-alternating design of the ablation ladder in the main paper; the primary method is the alternating variant, which applies the coupling MLP per head and contributes the $+0.33\%$ overhead reported in the caption and in the main text. GQA is included as a parameter-budget reference, not a quality baseline (see the main paper).

\begin{table}[h]
\centering
\caption{Measured wall-clock cost at 60M on one H100, batch 8, forward and backward. Throughput is tokens per second relative to standard attention (higher is better); peak memory is at sequence length 512. Because the per-head coupling MLPs are unfused, the realized cost exceeds the $0.33\%$ parameter overhead. At length 2048 Differential Attention is itself the costlier component.}
\label{tab:throughput}
\resizebox{\columnwidth}{!}{%
\begin{tabular}{lcccc}
\toprule
\textbf{Variant} & \textbf{Params} & \textbf{Rel.\ tok/s (512)} & \textbf{Rel.\ tok/s (2048)} & \textbf{Peak mem (MB)} \\
\midrule
Standard                     & $+0.00\%$ & $1.00$ & $1.00$ & $4175$ \\
Differential Attention       & $+0.00\%$ & $0.86$ & $0.68$ & $5199$ \\
Alternating coupling         & $+0.33\%$ & $0.90$ & $0.85$ & $5071$ \\
\quad ${+}$ Diff.\ Attention & $+0.33\%$ & $0.79$ & $0.61$ & $6095$ \\
\bottomrule
\end{tabular}
}
\end{table}

\paragraph{Cost at the scale of the main result.}
Table~\ref{tab:throughput455} repeats the measurement at 455M, the scale of
the central composition claim, at sequence lengths 512 (batch 8) and 2048
(batch 2), with batch fixed within each length so relative overheads are
directly comparable across variants.
The parameter overhead of the coupling falls to $0.13\%$ at 455M, and the
relative training-throughput overhead of the composition over Differential
Attention alone falls with both scale and length: about $7.4\%$ at length 512
and $4.1\%$ at length 2048, against $8$ to $10\%$ at 60M.
The coupling's cost is linear in sequence length and model width while
attention scoring is quadratic in length, so the coupling's share of total
compute shrinks exactly where the main result lives.
Peak-memory overhead follows the same logic: it comes from storing the
coupling activations, is largest in relative terms at the short length
(about $+23\%$ over Differential Attention at 512), and falls to about
$+11\%$ at length 2048, where attention's own sequence-quadratic buffers
dominate.

\begin{table}[h]
\centering
\caption{Measured wall-clock cost at 455M on one H100, forward and backward,
batch 8 at sequence length 512 and batch 2 at length 2048. Throughput is
tokens per second relative to standard attention at the same length; peak
memory is reported at both lengths. The composition's overhead over
Differential Attention alone is $7.4\%$ at length 512 and $4.1\%$ at length
2048.}
\label{tab:throughput455}
\resizebox{\columnwidth}{!}{%
\begin{tabular}{lccccc}
\toprule
\textbf{Variant} & \textbf{Params} & \textbf{Rel.\ tok/s (512)} & \textbf{Rel.\ tok/s (2048)} & \textbf{Mem 512 (GB)} & \textbf{Mem 2048 (GB)} \\
\midrule
Standard                     & $+0.00\%$ & $1.00$ & $1.00$ & $16.7$ & $26.4$ \\
Differential Attention       & $+0.00\%$ & $0.87$ & $0.72$ & $22.8$ & $51.3$ \\
Alternating coupling         & $+0.13\%$ & $0.91$ & $0.94$ & $22.1$ & $31.8$ \\
\quad ${+}$ Diff.\ Attention & $+0.13\%$ & $0.81$ & $0.69$ & $28.2$ & $56.7$ \\
\bottomrule
\end{tabular}
}
\end{table}

%% file: sections/supp_perseed.tex
\section{Per-Seed Perplexities and Effect Sizes}
\label{app:perseed}

This section lists the per-seed perplexities behind the per-comparison statistics in the main paper, for the comparisons whose six-seed headline disclosure is not already given there.

\paragraph{455M breadth comparisons (three seeds; seeds 42, 137, 2024).}
The per-seed perplexities are $19.2541$, $19.2791$, $19.1853$ for Multi-Token Attention and $19.1607$, $19.2168$, $19.2244$ for its composition ($p{=}0.16$), and $19.8369$, $19.8309$, $19.8026$ for query-key normalization and $19.6468$, $19.7372$, $19.8017$ for its composition ($p{=}0.08$); each composition has the lower mean but with overlapping seeds, consistent with non-significance at three seeds.

\paragraph{Rotary length sweep (three seeds each).}
For the sequence-length-1024 rotary comparison the per-seed values are $19.4719, 19.5286, 19.4854$ for Differential Attention and $19.1643, 19.2573, 19.1656$ for the composition, which separate without overlap.
The two additional rotary lengths of the sweep, counted in the Bonferroni family of the main paper but not surviving the correction, have the following per-seed values: at 512, $18.8498$, $18.8540$, $18.8541$ for Differential Attention and $18.6396$, $18.6897$, $18.6376$ for the composition (one-sided Welch $p{=}0.0035$), and at 2048, $20.4355$, $20.1022$, $20.2140$ and $19.8902$, $19.9353$, $19.6748$ ($p{=}0.016$); both lengths separate without overlap.
With three seeds these $p$-values and their effect sizes are unstable, and the 512 Differential Attention seeds cluster unusually tightly, so we read the sweep as directional confirmation rather than as precise estimates.

\paragraph{Effect sizes.}
Cohen's $d$ is reported in the main paper for completeness; with three seeds its confidence interval is wide, so for the three-seed rows the magnitudes indicate direction and separation rather than a stable effect-size estimate.
The headline 455M Differential comparison uses six seeds and gives $d{=}2.1$; with only three seeds this estimate is inflated by an unusually tight Differential Attention spread, so we report the more reliable six-seed value.

%% file: sections/supp_testppl.tex
\section{Held-Out Test-Set Perplexity}
\label{app:testppl}
The validation-selected 455M checkpoints, evaluated on the WikiText-103 test split (token-weighted over the full split, same tokenizer), reproduce the validation comparisons in ordering and significance (Table~\ref{tab:testppl}): the Differential composition is significant at sequence length 512 ($20.88$ against $21.02$, six seeds, $p{=}0.002$) and at sequence length 1024 with rotary embeddings ($19.81$ against $20.13$, three seeds, $p{=}0.003$), and the standalone comparison remains not significant ($21.76$ against $21.79$, $p{=}0.21$).
We treat these as replications of the corresponding validation tests on a held-out split, not as additional members of the Bonferroni family.
A per-position profile of the test loss shows the composition's gain over Differential Attention is positive in all twenty-four 64-token bins across both sequence lengths, rather than concentrated at particular positions.

\begin{table}[h]
\small
\centering
\caption{WikiText-103 test-split perplexity of the validation-selected 455M checkpoints, mean$\pm$std. Ordering and significance match the validation comparisons of the main paper's composition table, including the non-significant standalone gap.}
\label{tab:testppl}
\begin{tabular}{llrrr}
\toprule
\textbf{Comparison} & $n$ & \textbf{Baseline} & \textbf{Treatment} & $p$ \\
\midrule
Comp(Diff), seq 512 & 6 & $21.02 \pm 0.04$ & $\mathbf{20.88 \pm 0.07}$ & $0.002$ \\
Comp(Diff), seq 1024, RoPE & 3 & $20.13 \pm 0.08$ & $\mathbf{19.81 \pm 0.06}$ & $0.003$ \\
Standalone, seq 512 & 3 & $21.79 \pm 0.05$ & $21.76 \pm 0.02$ & $0.21$ \\
\bottomrule
\end{tabular}
\end{table}

%% file: sections/supp_analysis.tex
\section{Additional Analysis}
\label{app:additional_figures}

\paragraph{Attention matrix rank.}
\label{app:attn_rank}
We probe post-softmax and pre-softmax effective rank in the 455M (24-layer) models, seed 42,
averaged over validation sequences.
Effective rank is $(\sum_i \sigma_i)^2 / \sum_i \sigma_i^2$ where $\sigma_i$ are the singular
values of the attention map $A = \mathrm{softmax}(QK^\top/\sqrt{d_k})$ (post-softmax)
or of $QK^\top/\sqrt{d_k}$ (pre-softmax).

\textbf{Pre-softmax logit rank.}
The effective rank of the pre-softmax logit matrix is essentially identical across all variants
(approximately 14.4 for all four), because the causal mask dominates the singular-value spectrum.
Coupling does not change it.

\textbf{Post-softmax rank in late layers.}
Table~\ref{tab:attn_rank} reports mean effective rank over the last third of layers (L16--L23)
of the 455M model.

\begin{table}[h]
\centering
\caption{Post-softmax attention-map effective rank, mean over L16--L23 of the 455M (24-layer)
  model (seed 42, WikiText-103). Pre-softmax logit rank is $\approx 14.4$ for all variants
  (causal-mask dominated).}
\label{tab:attn_rank}
\begin{tabular}{lc}
\toprule
\textbf{Variant} & \textbf{Post-softmax eff.\ rank (L16--L23 mean)} \\
\midrule
Standard                              & 46.2 \\
Alternating coupling                  & 53.5 \\
Differential Attention                & 42.5 \\
\quad ${+}$ coupling (composition)    & 45.3 \\
\bottomrule
\end{tabular}
\end{table}

The standalone alternating coupling preserves modestly more post-softmax rank in late layers
relative to standard attention (53.5 vs.\ 46.2).
Differential Attention alone is \emph{lower} (42.5), and the composition partially recovers it
(45.3).
We interpret this as a modest effect: coupling shifts the rank distribution in late layers but
does not escape a rank-collapse bound.
In particular, the pre-softmax picture is the same for all variants, confirming that the
coupling's effect is downstream of the scoring step.

\paragraph{RoPE compatibility.}
\label{app:rope}
We verify that the alternating affine coupling composes with Rotary Position Embeddings (RoPE).
Table~\ref{tab:rope_ablation} shows 60M results for the alternating affine coupling
(the primary result in this paper) and standard attention, each trained with learned positional
embeddings and with RoPE.
Both variants see higher perplexity under RoPE at 60M scale; the coupling advantage is
preserved under both positional encoding choices.

\begin{table}[h]
\centering
\caption{RoPE ablation at 60M scale (alternating affine coupling). Mean $\pm$ std over 3 seeds.}
\label{tab:rope_ablation}
\begin{tabular}{lcc}
\toprule
\textbf{Attention} & \textbf{Learned Pos.} & \textbf{RoPE} \\
\midrule
Standard             & 24.20 {\scriptsize $\pm$ 0.13} & 25.84 {\scriptsize $\pm$ 0.07} \\
Alternating coupling & 23.92 {\scriptsize $\pm$ 0.04} & 24.00 {\scriptsize $\pm$ 0.03} \\
\bottomrule
\end{tabular}
\end{table}

\noindent The table reports both positional-encoding conditions (learned and RoPE)
for standard attention and for the alternating coupling.
The comparison isolates the effect of replacing learned positional embeddings with RoPE; the
relative ordering, with the coupling advantage preserved under both choices, is the claim, not
the absolute perplexities.
Standard attention degrades more under RoPE at this scale ($24.20$ to $25.84$) than the coupling
does ($23.92$ to $24.00$). We report this asymmetry without interpreting it: the setup is not
tuned for RoPE, and a single 60M comparison cannot separate a genuine interaction from a
configuration artifact.
The 455M longer-context runs in the main paper also use rotary embeddings, at sequence
lengths 512, 1024, and 2048; the composition gain across that sweep extends RoPE compatibility from this
standalone 60M check to the Differential Attention composition at scale.

%% file: sections/supp_optstability.tex
\section{Optimization Stability}
\label{app:hessian}

\paragraph{Cross-seed variance.}
Beyond mean perplexity, the alternating coupling produces the tightest seed spread of any
variant at 60M: its standard deviation over three seeds is $0.04$ PPL against $0.13$ for
standard attention (the statistical tests in the main paper), a threefold reduction, and the
Differential composition is likewise tight ($0.02$ against $0.08$ for Differential Attention at 150M).
This tightening is not uniform: at 150M the standalone spread is no smaller than standard,
so we report the variance reduction as an observation at specific scales rather than a
guaranteed effect.
Three-seed spreads are themselves noisy estimates of variance: at 455M, a Differential
Attention spread that three seeds suggested was unusually tight widened once six seeds were
available (main paper), and we therefore do not base any interpretation on seed spread alone.
Per-layer attention entropy profiles (Figure~\ref{fig:entropy_comparison}) provide descriptive
context and show no entropy effect that would explain the variance reduction.

\paragraph{Interpretation.}
Where the spread tightens, it is consistent with the coupling guiding optimization toward a
flatter basin, in which a smaller top Hessian eigenvalue would make the minimum less sensitive
to initialization.
We report the variance numbers as the direct measurement we have; a Hessian-eigenvalue probe,
using Pearlmutter products with power iteration, is a natural confirmation we leave to future
work.
We do not claim a flatness result beyond the variance observations.

%% file: sections/supp_entropy.tex
\section{Attention Entropy Profiles}
\label{app:entropy_profiles}
Attention entropy $S = -\sum_j a_j \log a_j$ measures distributional spread. Entropy collapse, in which heads concentrate on near-deterministic distributions, is associated with reduced representational capacity~\citep{zhai2023entropy}. Figure~\ref{fig:entropy_comparison} compares per-layer entropy across variants.

\begin{figure}[h]
\centering
\begin{minipage}[t]{0.49\textwidth}
\centering
\begin{tikzpicture}
\begin{axis}[
  width=\linewidth, height=0.8\linewidth,
  xlabel={Layer}, ylabel={Mean entropy (nats)},
  xmin=-0.3, xmax=7.3, ymin=1.5, ymax=5.7,
  legend style={font=\scriptsize, at={(0.98,0.98)}, anchor=north east, draw=none, fill=none},
  tick label style={font=\footnotesize}, label style={font=\small},
  ymajorgrids=true, grid style={gray!15},
]
\addplot[color=black, mark=*, mark size=1.2pt, thick] coordinates {(0,5.17)(1,2.51)(2,2.36)(3,2.57)(4,4.07)(5,3.30)(6,3.51)(7,4.00)};
\addlegendentry{Standard}
\addplot[color=orange!85!black, mark=triangle*, mark size=1.6pt, thick] coordinates {(0,5.21)(1,4.77)(2,3.25)(3,2.35)(4,2.58)(5,3.73)(6,4.25)(7,4.55)};
\addlegendentry{Differential Attention}
\addplot[color=blue!70!black, mark=square*, mark size=1.2pt, thick] coordinates {(0,5.02)(1,2.48)(2,2.23)(3,2.73)(4,3.43)(5,3.28)(6,3.40)(7,3.76)};
\addlegendentry{Coupling}
\end{axis}
\end{tikzpicture}
\\[2pt]
{\footnotesize (a)~60M}
\end{minipage}
\hfill
\begin{minipage}[t]{0.49\textwidth}
\centering
\begin{tikzpicture}
\begin{axis}[
  width=\linewidth, height=0.8\linewidth,
  xlabel={Layer}, ylabel={Mean entropy (nats)},
  xmin=-0.3, xmax=11.3, ymin=1.5, ymax=5.7,
  tick label style={font=\footnotesize}, label style={font=\small},
  ymajorgrids=true, grid style={gray!15},
]
\addplot[color=black, mark=*, mark size=1.0pt, thick] coordinates {(0,5.16)(1,2.97)(2,1.98)(3,2.25)(4,3.47)(5,4.16)(6,3.92)(7,3.59)(8,3.56)(9,3.75)(10,3.88)(11,3.97)};
\addplot[color=orange!85!black, mark=triangle*, mark size=1.4pt, thick] coordinates {(0,5.23)(1,4.16)(2,3.30)(3,2.59)(4,3.36)(5,4.01)(6,3.73)(7,4.08)(8,4.11)(9,4.33)(10,4.55)(11,4.63)};
\addplot[color=blue!70!black, mark=square*, mark size=1.0pt, thick] coordinates {(0,5.08)(1,2.54)(2,1.75)(3,2.54)(4,3.64)(5,3.97)(6,3.56)(7,3.61)(8,3.32)(9,3.43)(10,3.52)(11,3.82)};
\end{axis}
\end{tikzpicture}
\\[2pt]
{\footnotesize (b)~150M}
\end{minipage}
\caption{Per-layer mean attention entropy at the end of training (seed~42, mean over heads). All variants dip at the early attention-sink layer (Layer~2). The alternating coupling tracks standard attention closely and is marginally sharper (lower entropy) at the sink layer at both sizes, while Differential Attention holds higher entropy through the middle and late layers. We show these profiles as descriptive context; the stability evidence we rely on is the cross-seed variance reported in the main paper, not an entropy effect.}
\label{fig:entropy_comparison}
\end{figure}

%% file: sections/supp_hamiltonian.tex
\section{Physics Motivation: From Hamiltonian Mechanics to Coupled Dynamics}
\label{app:hamiltonian}

Coupled query-key dynamics was originally inspired by Hamiltonian mechanics; we record the path briefly because it explains the term \emph{dynamics} and why the final method is an affine coupling rather than a symplectic integrator.
We treated queries $Q$ as generalized position and keys $K$ as generalized momentum with separable energy $H(q, k) = \tfrac{1}{2}\|k\|^2 + V(q)$, so that Hamilton's equations
\begin{equation}
\dot{q} = \frac{\partial H}{\partial k} = k, \qquad \dot{k} = -\frac{\partial H}{\partial q} = -\nabla_q V(q)
\end{equation}
make $Q$'s update depend on $K$ and $K$'s update depend on a learned function of $Q$, the indirect coupling described in the main paper.
A leapfrog (St\"{o}rmer--Verlet) discretization of this flow is a composition of unit-Jacobian shear maps, so by Liouville's theorem it preserves phase-space volume~\citep{arnold1989mathematical}; in our prototypes the force was amortized by a learned MLP in the style of Hamiltonian neural networks~\citep{greydanus2019hamiltonian}.
That volume preservation is exactly the property the final method gives up: a purely additive, volume-preserving coupling is provably restricted, whereas the affine, non-volume-preserving, alternating coupling described in the main paper removes that restriction~\citep{teshima2020coupling}, and the ablation in the main paper, not this analogy, is what supports the design.